\pdfoutput=1 

\documentclass[nohyperref]{article}

\usepackage{microtype}
\usepackage{graphicx}
\usepackage{subfigure}
\usepackage{multirow}

\usepackage{booktabs} 
\usepackage{bbm}

\usepackage{url}

\usepackage{hyperref}

\usepackage[accepted]{icml2022}

\usepackage{amsmath}
\usepackage{amssymb}
\usepackage{comment}
\usepackage{mathtools}
\usepackage{amsthm}

\newtheorem{defn}{Definition}

\usepackage{amsmath,amsfonts,bm}

\def\eqref#1{equation~\ref{#1}}

\def\1{\bm{1}}

\DeclareMathAlphabet{\mathsfit}{\encodingdefault}{\sfdefault}{m}{sl}
\SetMathAlphabet{\mathsfit}{bold}{\encodingdefault}{\sfdefault}{bx}{n}

\newcommand{\R}{\mathbb{R}}

\newcommand{\sigmoid}{\sigma}

\DeclareMathOperator*{\argmin}{arg\,min}

\usepackage[capitalize,noabbrev]{cleveref}

\theoremstyle{plain}

\theoremstyle{definition}

\theoremstyle{remark}

\usepackage[textsize=tiny]{todonotes}

\icmltitlerunning{Zero-Shot AutoML with Pretrained Models}

\begin{document}

\twocolumn[
\icmltitle{Zero-Shot AutoML with Pretrained Models}



\icmlsetsymbol{equal}{*}

\begin{icmlauthorlist}
\icmlauthor{Ekrem \"Ozt\"urk}{equal,fr}
\icmlauthor{Fabio Ferreira}{equal,fr}
\icmlauthor{Hadi S. Jomaa}{equal,hil}
\icmlauthor{Lars Schmidt-Thieme}{hil}
\icmlauthor{Josif Grabocka}{fr}
\icmlauthor{Frank Hutter}{fr,bosch}
\end{icmlauthorlist}

\icmlaffiliation{fr}{University of Freiburg}
\icmlaffiliation{hil}{University of Hildesheim}
\icmlaffiliation{bosch}{Bosch Center for Artificial Intelligence}

\icmlcorrespondingauthor{Fabio Ferreira}{ferreira@cs.uni-freiburg.de}

\icmlkeywords{Machine Learning, ICML}

\vskip 0.3in
]



\printAffiliationsAndNotice{\icmlEqualContribution} 

\begin{abstract}
Given a new dataset $D$ and a low compute budget, how should we choose a pre-trained model to fine-tune to $D$, and set the fine-tuning hyperparameters without risking overfitting, particularly if $D$ is small? Here, we extend automated machine learning (AutoML) to best make these choices. Our domain-independent meta-learning approach learns a zero-shot surrogate model, which, at test time, allows to select the right deep learning (DL) pipeline (including the pre-trained model and fine-tuning hyperparameters) for a new dataset $D$ given only trivial meta-features describing $D$, such as image resolution or the number of classes. To train this zero-shot model, we collect performance data for many DL pipelines on a large collection of datasets and meta-train on this data to minimize a pairwise ranking objective. We evaluate our approach under the strict time limit of the vision track of the ChaLearn AutoDL challenge benchmark, clearly outperforming all challenge contenders. 


\end{abstract}

\section{Introduction}
\label{sec:intro}
A typical problem in deep learning (DL) applications is to find a good model for a given dataset $D$ in a restrictive time budget. 
In the case of tabular data, a popular approach for solving this problem is automated machine learning (AutoML), as implemented, e.g., in Auto-sklearn~\cite{feurer-nips2015a} or Auto-Gluon~\cite{erickson-arxiv20a}. However, in domains such as computer vision and natural language processing, a better solution, especially under low resource constraints, is typically to fine-tune an existing pre-trained model.
This, at first glance, appears to render AutoML unnecessary for those domains. However, as we will demonstrate in this paper, AutoML and pre-trained models can be combined to yield much stronger performance than either of them alone.

A great advantage of fine-tuning pre-trained models is strong anytime performance: the use of pre-trained models often allows to obtain very strong performance orders of magnitude faster than when training a model from scratch.
In many practical applications, this strong anytime performance is crucial, e.g., for DL-based recognition systems in manufacturing, or automated DL (AutoDL) web services. The clock starts ticking as soon as a new dataset is available, and it would be far too costly to train a new model from scratch, let alone optimize its hyperparameters.
The recent ChaLearn AutoDL competition~\citep{liu-tpami21a} mimicked these tight time constraints, rewarding performance found in an anytime fashion.

While fine-tuning pre-trained models enjoys strong anytime performance, it also introduces many additional degrees of freedom. Firstly, we need to select a pre-trained network to fine-tune. To obtain good anytime performance, we may even want to start by training a shallow model to obtain good results quickly, and at some point switch to fine-tuning a deeper model. There are many additional degrees of freedom in this fine-tuning phase, concerning learning rates, data augmentation, and regularization techniques. We refer to the combination of a pre-trained model and the fully specified fine-tuning phase, including its hyperparameters, as a \emph{DL pipeline}.
Which DL pipeline works best depends heavily on the dataset, for instance, datasets with high-resolution images may favor the use of more downsampling layers than the low-resolution images of the CIFAR dataset~\cite{cifar}; likewise, datasets with few images may favor smaller learning rates.
We, therefore, require an automated method for selecting the best DL pipeline based on the characteristics of the dataset at hand.

In this paper, we tackle this problem by meta-learning a model across datasets that enables zero-shot DL pipeline selection. Specifically, we create a meta-dataset holding the performance of many DL pipelines on a broad range of datasets. Using this meta-dataset, we then learn a function that selects the right DL pipeline based on the properties of the dataset (e.g., the image resolution and the number of images) in a zero-shot setting. %
To learn this selection function, we first formulate DL pipeline selection as a classical algorithm selection (AS) problem~\citep{rice76a} and then improve upon this formulation by recognizing DL pipelines as points in a geometric space that allows information about the performance of some pipelines to inform performance on others. We then train a deep neural network with a pairwise ranking objective to emphasize the rank of the DL pipeline predicted to perform best in a manner that automatically normalizes across datasets. Note, that we use the \emph{zero-shot} nomenclature not to refer to samples of unseen classes but to express that we cannot even afford a single exploratory evaluation of a pipeline but need to directly select a suitable one in a zero-shot manner.

Our contributions can be summarized as:
\begin{itemize}
\vspace*{-0.2cm}
    \item We extend AutoML to best exploit pre-trained models by meta-learning to select the best DL pipeline conditional on dataset meta-features.
\vspace*{-0.2cm}
    \item We introduce a large meta-dataset with the performances of 525 DL pipelines across 35 image-classification datasets and 15 augmentations each. With 525*35*15 entries, it is, to our best knowledge, the first DL meta-dataset for image classification of this size, being over 1000 times larger than previous meta-datasets~\cite{triantafillou-arxiv19a}.
\vspace*{-0.2cm}
    \item We go beyond a standard formulation as an algorithm selection problem by formulating the new problem of selecting a DL pipeline as a point in a geometric space to exploit similarities between DL pipelines.
\vspace*{-0.2cm}
    \item We introduce a novel zero-shot AutoDL method that addresses this pipeline selection problem with a pairwise ranking loss.
\vspace*{-0.2cm}
    \item In the setting of the recent ChaLearn AutoDL challenge~\citep{liu-tpami21a}, our zero-shot AutoDL approach dominates all competitors on a broad range of 35 image datasets, as well as in the challenge itself.
\end{itemize}

To foster reproducibility, we make our PyTorch \citep{pytorch-neurips19a} code, models, and data publicly available under \href{https://github.com/automl/zero-shot-automl-with-pretrained-models}{this URL}.

\section{Related Work}
\label{sec:relatedwork}

\paragraph{Algorithm selection}
Assume a set $\mathcal{P}$ of algorithms $\mathcal{A} \in \mathcal{P}$ (e.g., classifiers or neural network hyperparameter configurations), a set of \emph{instances} $i\in \mathcal{I}$ (e.g., dataset features), and a \emph{cost metric} $m:\mathcal{P} \times \mathcal{I}\to \mathbb {R}$. Specifying the loss of algorithm $\mathcal{A} \in \mathcal{P}$ on instance $i\in \mathcal{I}$ with $m(\mathcal{A},i)$, the algorithm selection problem~\citep{rice76a,smith-cs08,kotthoff-aicom12a,bischl-aij16a} is to find a mapping $s:\mathcal{I}\to \mathcal{P}$ that minimizes the cost metric $\sum_{i\in {\mathcal {I}}}m(s(i),i)$ across instances $\mathcal{I}$. Algorithm selection has been applied to achieve state-of-the-art results in many hard combinatorial problems, most prominently Boolean satisfiability solving (SAT), where SATzilla~\citep{xu-jair08a} won several competitions by learning to select the best SAT solver on a per-instance basis. There are many methods for solving algorithm selection, based on regression~\citep{xu-jair08a}, k-nearest neighbours~\citep{kadioglu-cp11a}, cost-sensitive classification~\citep{xu-sat12a}, and clustering~\citep{kadioglu-ecai10,malitsky-ijcai13a}. \emph{AutoFolio}~\citep{lindauer-aaai15a} is a state-of-the-art algorithm selection system that combines all of these approaches in one and chooses between them using algorithm configuration~\citep{hutter-lion11a}. We will use AutoFolio as one of our methods for selecting DL pipelines based on dataset meta-features.

\paragraph{Hyperparameter optimization (HPO)}
HPO plays an integral role in fine-tuning any machine learning algorithm. Beyond simple strategies, such as random search~\cite{bergstra-jmlr12a}, conventional techniques typically involve fitting (probabilistic) surrogate models of the true response, e.g. Gaussian Process~\cite{rasmussen-book06a}, random forests~\cite{hutter-lion11a}, neural networks~\cite{springenberg-nips2016}, or some hybrid techniques~\cite{snoek-icml15a}, and selecting configurations that optimize pre-defined acquisition functions~\citep{wilson-neurips18a}. Recently, approaches started taking into account the dissimilarity between pre-training and downstream domains~\citep{li-iclr2020a}. HPO multi-fidelity methods further reduce the wall-clock time necessary to arrive at optimal configurations~\cite{li-iclr17a,falkner-icml18a,awad-ijcai21a}.

\paragraph{Transfer HPO} 
Transfer learning can be used in HPO to leverage knowledge from previous experiments to yield a strong surrogate model with few observations on the target dataset. For example, \citealt{wistuba-iclr21a} and \citealt{Jomaa2021_DKLM} both propose a meta-initialization strategy by optimizing a deep kernel Gaussian process surrogate model~\cite{wislon-aistats16a} across meta-train datasets to allow for fast adaptation given a few observations. Similarly, \citealt{salinas-icml20a} learns a Gaussian Copula~\cite{wilson-neurips10a} and addresses the heterogeneous scales of the responses across datasets, whereas \citealt{perrone-neurips18a} pre-trains a shared layer in a multi-task setting.
Transfer HPO is also possible based on meta-features~\cite{vanschoren-arxiv18a}, i.e. dataset characteristics which can be either engineered~\cite{feurer-aaai15a,wistuba-ecml16a} or learned~\cite{jomaa-dmkd21a} to warm-start HPO.

\paragraph{Zero-shot HPO} 
The conventional setting of zero-shot learning aims to recognize samples whose instances may not have been seen during training~\cite{xian-cvpr18a, verma-cvpr18a, radford-icml21a}. In the setting of zero-shot HPO, in contrast, the focus lies on improving sample efficiency for hyperparameter optimization.
Contrary to techniques in previous sections that improve their sample efficiency by increasing the number of trials, zero-shot HPO has emerged as a more efficient approach that does not require any observations of the response on the target dataset. \citealt{wistuba-icdm15} design a sequential model-free approach that optimizes the ranks of hyperparameter configurations based on their average performance over a collection of datasets. \citealt{winkelmolen-arxiv20a} propose a Bayesian optimization solution for zero-shot HPO, whereby a surrogate model is fit to the dataset and hyperparameters and optimized by minimizing a ranking meta-loss. We note that both these approaches return a fixed set of hyperparameter configurations without using meta-features, which is undesirable as the AutoDL setting used in this work only allows for running a single model.
Related to our work is \citep{tornede-ds20a}, who also describe datasets and pipelines as joint feature vectors. They use these to assess the learning of zero-shot models with algorithm selection and ranking-based objectives in a benchmark of tabular datasets and shallow base models in a sparse cost-matrix setting. In this paper, we propose a novel zero-shot HPO solution inspired by the success of algorithm selection techniques that learns to select the best DL pipeline based on both parametric choices inside the DL pipeline and dataset meta-features of complex vision datasets, by optimizing a ranking objective jointly across datasets. 
\paragraph{AutoDL Competition} 
ChaLearn’s AutoDL Challenge~\cite{liu-tpami21a} evaluated competitors in an anytime setting with strict time limits, leading to the prominent use of pre-trained models by the participants. We focus on the challenge's image-track and summarize the winning approaches here and give more details in Section \ref{exp: baselines}.
In the 2019 AutoCV/CV2 competition, the winning approach \citep{baek-arxiv20a} used a ResNet-18~\cite{he-corr15a} pre-trained on ImageNet~\cite{krizhevsky-nips12} with Fast AutoAugment~\cite{lim-neurips19a}. All image-track winning solutions used the AutoCV winner code as a skeleton and built their methods on top. Their modifications ranged from switching to a more stable ResNet-9 during training (DeepWisdom), ensembling predictions (DeepBlueAI) to data-adaptive pre-processing (PASA NJU). 

\paragraph{Meta-learning} Meta-learning~\cite{finn-icml17a} can be used to solve tasks where the training dataset is small. \citealt{sun-cvpr19a} meta-learn to transfer large-scale pre-trained DNN weights to solve few-shot learning tasks. \citealt{verma-aaai19a} tackle Zero-Shot Learning by meta-learning a generative model for synthesizing examples from unseen classes conditioned on class attributes. \citealt{laadan2019rankml} generate (shallow model) pipelines on diverse datasets and use dataset meta-features to rank the pipelines to create a meta-dataset of pipelines and their performance results.

Despite the abundance of meta-learning methods, and in contrast to the large benchmarks for tabular data~\cite{arango-arxiv21a}, few meta-learning benchmarks exist for image datasets. \citealt{zhai-corr19a} introduced a set of 19 vision tasks and evaluated 18 representation learning methods. \citealt{triantafillou-arxiv19a} also introduced a meta-dataset of 10 few-shot image tasks and a growing set of baselines, currently comprising 11 and 18 evaluations on two different settings. ~\citealt{dumoulin-corr21a} combines these two benchmarks and compares Big Transfer~\cite{kolesnikov-eccv20a} against the baselines of \citealt{triantafillou-arxiv19a}. As we will show, our DL meta-dataset for image tasks is far larger than all previous meta-learning benchmarks.
\section{Zero-Shot AutoML with Pretained Models}
\label{sec:method}

\subsection{Notation and Problem Definition} 
Let $\mathcal{X} := \left\{x_n\right\}_{n=1}^N$ denote a set of $N$ distinct deep learning (DL) pipelines. Every DL pipeline $x_n:=\left(M_n,\theta_n\right)$ comprises a pre-trained model $M_n \in \mathcal{M}$ and fine-tuning hyperparameters $\theta_n \in \Theta$ that are used to fine-tune $M_n$ to a given dataset. Furthermore, let $\mathcal{D} = \{D_{i}\}_{i=1}^I$ denote a collection of $I$ datasets, where each dataset $D_i \in \mathcal{D}$ is split into disjoint training, validation and testing subsets $D_{i}:= D_i^\text{(tr)} \cup D_i^\text{(val)} \cup D_i^\text{(test)}$. For each dataset, we are given a vector of $K$ descriptive characteristics (a.k.a. meta-features), such as the number of data points and the image resolution, as $\phi_i \in \Phi \subseteq \mathbb{R}^K$ (see Section \ref{subsec:datasets} for the full set of meta-features we used in our experiments). 
We denote by $x_n^{(\text{ft})} := \text{Tune}\left(x_n, D^\text{(tr)}, D^\text{(val)}\right)$ the model resulting from fine-tuning the pre-trained model $M_n$ with hyperparameters $\theta_n$ on training data $D^\text{(tr)}$ using validation data $D^\text{(val)}$ for early stopping.
Then, denoting the loss of a fine-tuned model $x_n^{(\text{ft})}$ on the test split of the same dataset $D$ as $\mathcal{L}(x_n^{(\text{ft})}, D^\text{(\text{test})})$, the test cost of DL pipeline $x_n$ on $D$ is defined as:
\begin{equation}
\label{eq:cost} C(x_n,D) = \mathcal{L}\left(\text{Tune}\left(x_n, D^\text{(tr)}, D^\text{(val)}\right), D^\text{(test)}\right).
\end{equation}

\begin{defn}\label{def:ZeroShotAutoML}
Given a set of $N$ DL pipelines $\mathcal{X} := \left\{x_n\right\}_{n=1}^N$ and a collection of $I$ datasets $\mathcal{D} = \{D_{i}\}_{i=1}^I$ with meta-features $\phi_i$ for dataset $D_i \in \mathcal{D}$, and a $N \times I$ matrix of costs $C\left(x_n, D_i\right)$ representing the cost of pipeline $x_n$ on dataset $D_i$, the problem of \textbf{zero-shot AutoML with pre-trained models (ZAP)} is to find a mapping $f: \Phi \to \mathcal{X}$ that yields minimal expected cost over $\mathcal{D}$:
\begin{equation}
\label{eq:exp_cost} 
\operatorname{argmin}_f \; \mathbb{E}_{i\sim \{1,\dots,I\}} \left[ C( f(\phi_i), D_i ) \right].
\end{equation}
\end{defn}

\subsection{ZAP via Algorithm Selection (ZAP-AS)}

The problem of zero-shot AutoML with pre-trained models from Definition \ref{def:ZeroShotAutoML} can be directly formulated as an algorithm selection problem: the DL pipelines $\mathcal{X} := \left\{x_n\right\}_{n=1}^N$ are the algorithms $\mathcal{P}$, the datasets $\{D_{i}\}_{i=1}^I$ are the instances $\mathcal{I}$, and the test cost $C(x_n,D)$ of DL pipeline $x_n$ on $D$ defines the cost metric $m:\mathcal{P}\times \mathcal{I} \to \mathbb{R}$.
We use the state-of-the-art algorithm selection system AutoFolio to learn a selector between our predefined DL pipelines, since it subsumes approaches based on regression, classification, clustering, and cost-sensitive classification, and selects the best one for the data at hand~\citep{lindauer-aaai15a}. 

While this formulation of zero-shot AutoML with pre-trained models as algorithm selection will turn out to already yield very strong performance, it has one limitation: algorithm selection abstracts away our DL pipelines as uncorrelated algorithms, losing all information about the pre-trained models they are based on, and which hyperparameters are being used for fine-tuning. This information, e.g., allows us to predict the cost of DL pipelines to other DL pipelines with similar settings without ever having run them. Thus, we next introduce a novel approach for exploiting this knowledge.

\subsection{ZAP via Zero-Shot HPO (ZAP-HPO)}
\begin{figure}[t]
\begin{centering}
    \includegraphics[scale=0.65]{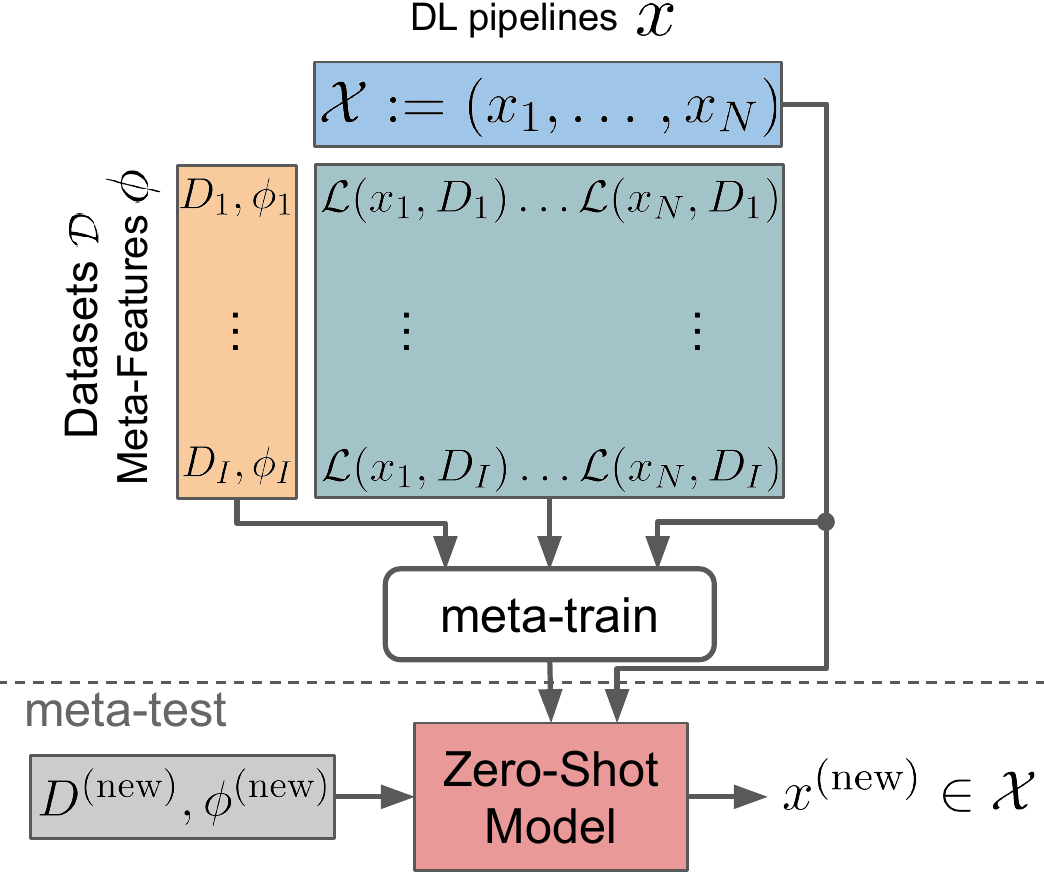}
    \caption[]{ZAP consists of two stages. In the meta-train stage, the cost matrix on the source tasks is leveraged to learn a joint response surface conditioned on the meta-features and pipelines. During the meta-test stage, ZAP assigns scores to the pipelines of the unseen datasets and the highest-scoring pipeline is selected.}
    \label{fig:overview}
\end{centering}
\vspace*{-0.3cm}
\end{figure}

We now describe a variant of our formulation of zero-shot AutoML that exploits the fact that the DL pipelines between which we select are points in a geometric space, and that we can see the space of DL pipelines we consider as a search space for hyperparameter optimization (HPO), with a categorical value for the choice of pre-trained model and continuous fine-tuning hyperparameters; we can then use concepts from zero-shot HPO to tackle this problem.

We define $M$ as a finite collection of $N$ pre-trained models and represent each instance, $M_n$, as a one-hot encoded vector, and $\theta_n\in\Theta\subseteq\mathbb{R}^L$ as a vector of continuous variables defining its respective hyperparameters. For instance, $\Theta$ can represent the continuous space of learning rates and dropout values of a pre-trained neural network model in $\mathcal{M}\in\{0,1\}^{|N|}$.
Consequently, the DL pipelines are projected to the geometric space defined by $\mathcal{X}\subseteq\mathcal{M}\times\Theta$ and can be viewed as a hyperparameter configuration space where pre-trained models are simply categorical variables. 

Denote by $f_\psi$ a parametric surrogate with parameters $\psi$ that estimates the test cost observed by fine-tuning the DL pipeline $x_j$ on dataset $D_i$ with meta-features $\phi_i$. The surrogate captures the fusion of (i) pipeline hyperparameters  (i.e. $x$ represented by the pre-trained model's one-hot-encoding indicator $\mathcal{M} \in \left\{1,\dots,M\right\}$ and the fine-tuning hyperparameters $\theta \in \Theta$ ) with (ii) dataset meta-features $\phi$, in order to estimate the cost after fine-tuning. Formally, that is:

\begin{align}
    \label{eq:surrogate}
    f\left(\psi\right)_{i,j} := f\left(x_j, \phi_i; \psi\right):\mathcal{M} \times \Theta \times \Phi \rightarrow \R_{+}
\end{align}

\begin{figure}[h!]
    \centering
    \includegraphics[width=0.9\columnwidth]{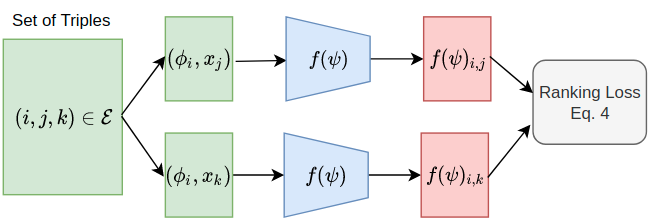}
    \caption{Overview of our pairwise ranking objective}
    \label{fig:flow}
\end{figure}

A unique aspect of searching for efficient pipelines is that we are concerned with the \textit{relative} cost of the pipelines, to find the best one. As such, we propose to utilize the surrogate model as a proxy function for the rank of configurations, and learn the pairwise cost ranking of pairs of pipelines. In this perspective, pairwise ranking strategies use the relative ordering between pairs of configurations to optimize the probability that the rank of the $j$-th pipeline is lower than the $k$-th pipeline on the $i$-th dataset. Therefore, using given pre-computed cost $C_{i,j}=C(x_j,D_i)$ we define the set of triples $\mathcal{E} := \left\{(i,j,k) \; | \; C\left(x_j, D_i\right) < C\left(x_k, D_i\right) \right\}$. Every triple $(i,j,k)$ denotes a pair $(x_j,x_k)$, where the cost of $x_j$ is smaller (better pipeline) than $x_k$ on the $i$-th dataset. Correspondingly, we want our surrogate to predict $f\left(\psi\right)_{i,j}$ to be lower than $f\left(\psi\right)_{i,k}$; we thus meta-learn our surrogate with a ranking loss as: 
\begin{equation}
\label{eq:ce}
    \argmin_\psi \sum_{(i, j, k)  \in \mathcal{E}}  \;  \log \left( \sigma\left( f\left(\psi\right)_{i,j} -  f\left(\psi\right)_{i,k}\right) \right),
\end{equation}

with $\sigmoid(\cdot)$ as the sigmoid function which prevents the difference from exploding to negative infinity as we minimize the loss. Equation~\ref{eq:ce} maximizes the gap between the surrogate scores, by \textit{decreasing the surrogate score for the good DL pipelines with low costs}, while at the same time increasing the surrogate score of bad pipelines with high costs. As a result, the score of the best DL pipeline with the lowest cost will be maximally decreased. Furthermore, Figure~\ref{fig:flow} presents a visual description of our proposed ranking loss.

Once we meta-learn the surrogate, we can transfer it to a new dataset $D^{(\text{new})}$ with meta-features $\phi^{(\text{new})}$ in a \textbf{zero-shot HPO} mechanism using Equation~\ref{eq:transfersurrogate}. The full meta-learn and meta-test procedure is depicted in Figure \ref{fig:overview}.

\begin{align}
    \label{eq:transfersurrogate}
    x^{(\text{new})} := \argmin\limits_{x_n, n\in\left\{1,\dots,N\right\}} f\left(x^{(\text{ft})}_{n}, \phi^{(\text{new})}; \psi\right)
\end{align}

For an empirical motivation on the benefits of learning surrogates with pairwise ranking losses, we compare to the same surrogate model optimized with a least-squares loss:

\begin{equation}
\label{eq: least-squared}
    \argmin_\psi \sum_{i=1}^I \sum_{n=1}^N  \; \left( f\left(\psi\right)_{i,n} -  C\left(x_n, D_i\right)\right) ^2
\end{equation}
 
\begin{figure}[h]
    \centering
    \includegraphics[width=1\columnwidth]{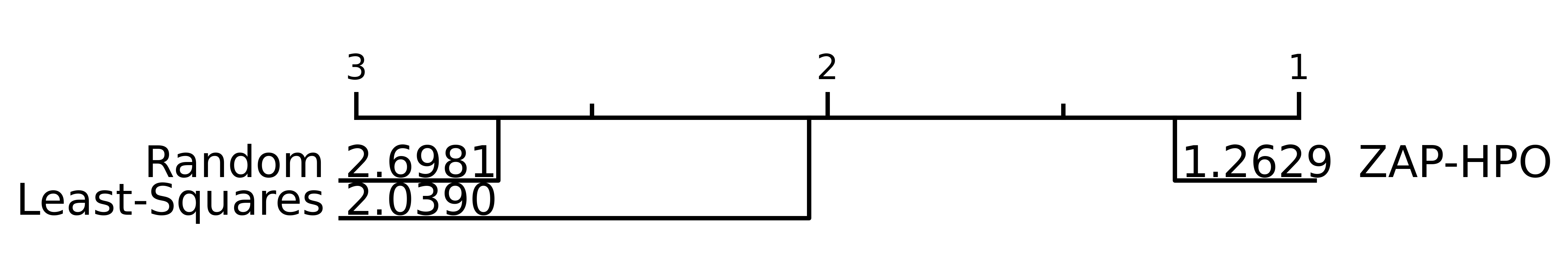}
\vspace*{-0.5cm}
    \caption{Critical difference diagram comparing loss functions using the Wilcoxon-Holm signed-rank (5\% significance level).}  
    \label{fig:critical}
\end{figure}

As a sanity check, we also compare the performance of randomly selecting a pipeline. In this experiment, we evaluate the performance of the pipeline having the largest estimated value by the surrogate (Equation~\ref{eq:transfersurrogate}) across all the $I$-many source datasets. The results of Figure~\ref{fig:critical} demonstrate that the surrogate trained with Equation~\ref{eq:ce} is significantly better than the regression-based variant of Equation~\ref{eq: least-squared} in terms of identifying the best pipeline. Further details about the evaluation protocol are found in Section \ref{subsec:eval_protocol}. 
\section{ZAP Meta-Dataset Design}
\label{sec:meta_datasets}
In this section, we introduce a novel meta-dataset~\cite{arango-arxiv21a}, that will ultimately allow us to perform zero-shot AutoML with pre-trained models (ZAP).
The meta-data required for the ZAP problem includes a set of datasets with meta-features, a set of DL pipelines, and the test costs for these pipelines on these datasets. Correspondingly, we describe how we curated a set of 35 image datasets, with 15 augmentations each (\ref{subsec:datasets}); define a space of DL pipelines (\ref{subsec:DL_pipeline_design_space}); and find strong instantiations in it for each of the datasets, each of which we evaluate on all datasets to obtain a $525\times 525$ matrix of test costs (\ref{subsec:perf-matrix}).

\subsection{A Set of Image Datasets for ZAP} 
\label{subsec:datasets}
The set of datasets should be chosen to be representative of the datasets that will eventually be tackled by the ZAP system building on them. While all our pre-trained networks are pre-trained on ImageNet~\cite{deng-cvpr09}, during the fine-tuning stage also smaller and specialized datasets are to be expected. Consequently, we chose both small and large, as well as diverse datasets that cover a wide range of domains (objects, medical, aerial, drawings, etc.) with varying formats, i.e. colored and black-and-white images and datasets with varying image resolutions or the number of classes. With this preference in mind, we retrieved 35 \emph{core} datasets provided by the TensorFlow~\citep{tensorflow} Datasets (TFDS) utility library~\citep{TFDS} and applied a dataset augmentation process~\citep{stoll2020_icgen} that takes a TFDS core dataset as input and outputs a subset of that differs in the number of classes and the number of train/test samples per class. Note that this dataset augmentation process does not perform augmentations on a sample level. We repeat this subset retrieval 15 times for each dataset, resulting in 525 datasets $\mathcal{D}$. Further details about the augmentation process are found in Appendix \ref{appendix:dataset_augmentation}. 

To represent a dataset, we use only extremely cheap and readily available dataset-dependent meta-features~\cite{vanschoren-automlbook20a} $\phi$: number of training images, number of image channels, image resolution, and number of classes.

\subsection{DL Pipeline Design Space for ZAP on Image Data} \label{subsec:DL_pipeline_design_space}
The DL pipelines we employ should be chosen to be diverse and achieve high performance on the aforementioned datasets since the optimum we can hope for is to choose the best of these pipelines on a per-dataset basis. To obtain strong pipelines, we started from the code base of the winner of the AutoCV competition~\citep{baek-arxiv20a}, which fine-tuned a pre-trained ResNet-18 model. We then built a highly-parameterized space of DL pipelines around this by exposing a wide range of degrees of freedom.
These included fine-tuning hyperparameters, such as learning rate, percentage of frozen parameters, weight decay, and batch size. Additionally, we exposed hyperparameters for the online execution that were previously hard-coded and that control, e.g., the number of samples used or when to evaluate progress with the validation dataset. To span a more powerful space with diverse pipelines, we also added additional architectural, optimization, as well as fine-tuning choices, including:

\begin{itemize}
    \item A binary choice between an EfficientNet~\cite{tan-icml19a} pre-trained on ImageNet~\cite{russakovsky-ijcv15} or the previously-used ResNet-18;
    \item The proportion of weights frozen when fine-tuning;
    \item Additional stochastic optimizers (Adam \cite{kingma-iclr15}, AdamW \cite{loshchilov-iclr18},  Nesterov accelerated gradient \cite{nesterov-1983}) and learning rate schedules (plateau, cosine \cite{loshchilov-iclr17a});
    \item A choice of using a simple classifier (either a SVM, random forest or logistic regression) that can be trained and used within the first 90 seconds of run-time in order to improve anytime performance.
\end{itemize}

Overall, our DL pipeline space $\mathcal{X}$ is comprised of 26 hyperparameters of the types real and integer-valued, categorical, and conditional. A condensed version is presented in Table~\ref{tab:ss_autodl1}.


\begin{table}[ht!]
    \small
    \caption[Table caption]{\textbf{The search space of our DL pipelines} consisting of general DL hyperparameters, training-strategy hyperparameters and fine-tuning strategy hyperparameters. A more detailed version can be found in Appendix \ref{appendix:dl_search_space}. }  
\begin{center}
    \begin{tabular}{lcc}
        \toprule
        \textbf{Name} & \textbf{Type, Scale} & \textbf{Range} \\ 
        \midrule
        Batch size & int, log & $[16, 64]$ \\ 
        Learning rate & float, log & $[10^{-5}, 10^{-1}]$ \\
        Weight decay & float, log & $[10^{-5}, 10^{-2}]$ \\
        Momentum & float & $[0.01, 0.99]$ \\
        Optimizer & cat & \{SGD, Adam,  \\
         &  & \text{   }AdamW\} \\
        Scheduler & cat & \{plateau, cosine\} \\
        Architecture & cat & \{ResNet18, EffNet-b0 \\
         &  & \text{   }EffNet-b1, EffNet-b2\} \\
        \hline
        Steps per epoch     & int, log & $[5, 250]$ \\
        Early epoch         & int& $[1, 3]$ \\
        CV ratio      & float & $[0.05, 0.2]$ \\
        Max valid count     & int, log & $[128, 512]$ \\
        Skip valid thresh. & float & $[0.7, 0.95]$ \\
        Test freq. & int & $[1, 3]$ \\
        Max inner loop & float & $[0.1, 0.3]$ \\
        \# init samples   & int,log & $[128, 512]$ \\
        Max input size            & int & $[5, 7]$ \\
        $1^{\text{st}}$ simple model & cat & \{true, false\} \\
        Simple model & cat & \{SVC, NuSVC, RF, LR\} \\
        \bottomrule
    \end{tabular}
    \end{center}
    \label{tab:ss_autodl1_excerpt}
\end{table}

\subsection{Selection and Evaluation of DL Pipelines}
\label{subsec:perf-matrix}

With the 525 datasets and our 26-dimensional DL pipeline space at our disposal, we now explain how we generated the DL pipeline candidates that we evaluated on the datasets. Instead of uniformly or randomly sampling the 26-dimensional DL pipeline space, to focus on DL pipelines that are strong at least on one dataset, we ran an optimization process to find a (near-)optimal DL pipeline for one dataset at a time. 
Specifically, we used the hyperparameter optimization method BOHB~\cite{falkner-icml18a}, which supports high-dimensional and categorical hyperparameter spaces, to find a (near)-optimal instantiation of our DL pipeline space for each dataset. We optimized the anytime Area under the Learning Curve (ALC) score (introduced in the AutoDL challenge~\cite{liu-tpami21a} and described in more detail in Section \ref{subsec:eval_protocol}) via BOHB, with a budget of five minutes for evaluating one DL pipeline on one dataset. We repeated each of these runs three times and used the mean to handle the substantial noise in these evaluations. This process resulted in one optimized DL pipeline per dataset; we thus have $N=D=525$ DL pipelines that comprise the set $\mathcal{X}$ of DL pipelines in our ZAP formulation.

Given this set of $525$ DL pipelines $\mathcal{X}$, and the set of our $525$ datasets $\mathcal{D}$, let us now explain the evaluation procedure. We ran each pipeline $x\in\mathcal{X}$ on each dataset $D\in\mathcal{D}$, computing the ALC score the pipelines reached within 10 minutes, and again computing the mean of three runs to reduce noise. While the AutoDL competition used a budget of 20 minutes, we used a shorter time of 10 minutes here (and 5 minutes for the runs of BOHB above) for two reasons: First, to limit the substantial computational overhead for carrying out these $525 \cdot 525=275,625$ evaluations of (DL pipeline, dataset) pairs; still, it required 2,871 GPU days to collect this data. Second, due to the typically monotonically increasing anytime ALC score, performance after 5 and 10 minutes can be expected to be a good proxy for the full 20 minutes.

\begin{figure}[ht]
\centering
    \includegraphics[trim={0 0 10cm 0}, clip, scale=0.07]{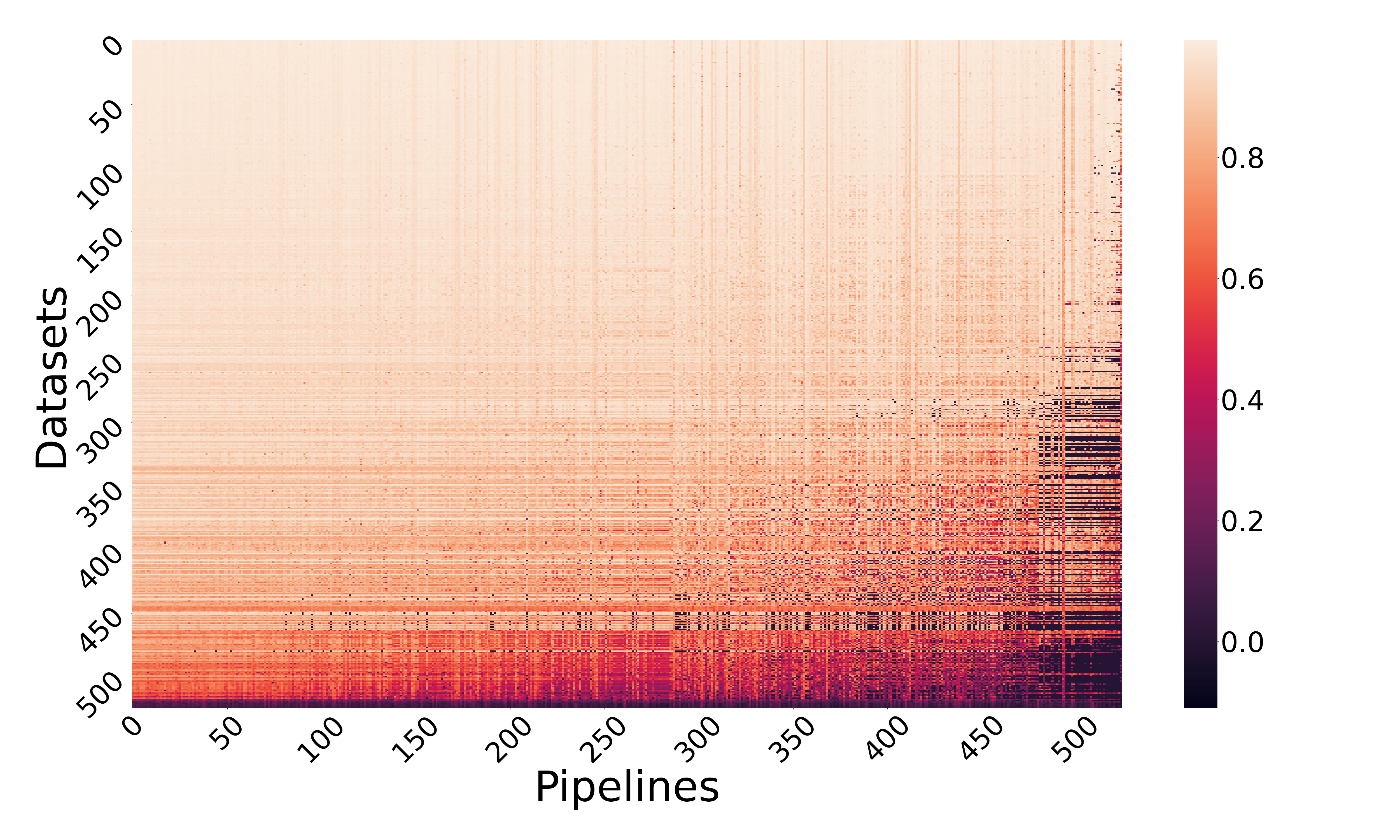}
    \vspace*{-0.5cm}
    \caption[]{\textbf{Cost matrix $\mathbf{C}$ as a heatmap} Color indicates the ALC score (higher is better). We observe that some datasets (dark rows) are more complex and some pipelines (dark columns) generalize worse across datasets systematically than others.}
    \label{fig:heatmap_perf_matrix}
\end{figure}

Finally, we record every pairs’ average-of-three ALC score in the cost matrix $C \in \R^{NxI}$ (in our case with $N=I=525$ since we found one DL pipeline per dataset). This cost matrix is visualized in Figure \ref{fig:heatmap_perf_matrix}. From the cost matrix, we directly see that there are easy datasets (at the top, where all pipelines achieve high scores) and hard ones at the bottom (where only very few pipelines reach high scores). Likewise, there are overall strong pipelines (to the left, with good scores on most datasets) and poor ones (on the right, with good scores on only a few datasets).
The most interesting pattern for ZAP is that there exists substantial horizontal and vertical striping, indicating that different datasets are hard for different pipelines. This points to the usefulness of selecting pipelines in a dataset-dependent manner in ZAP.

\section{Experiments}
\label{sec:experiments}
Our experiments are designed to evaluate the performance of an AutoDL system based on ZAP. We use the anytime evaluation protocol of the ChaLearn AutoDL Challenge~\cite{liu-tpami21a} and demonstrate that our ZAP methods perform substantially better than the winners of that competition, both in our ZAP evaluation benchmark grounded on our ZAP meta-dataset as well as in the original AutoDL challenge benchmark. We also carry out a range of ablations to better understand the root of our gains. 

\subsection{Evaluation Protocol}
\label{subsec:eval_protocol}
Let us first describe the restrictions of the evaluation protocol under which we evaluate different AutoDL methods (our ZAP approach and various baselines). The main restriction is the anytime learning metric to score participants with the Area under the Learning Curve (ALC): at each time step, an AutoDL system can update its predictions on test data, and in a post hoc evaluation phase, the accuracy of these predictions is averaged over the entire learning curve. A second core aspect of the challenge is the limited time budget of 20 minutes for training models and making predictions on test data. 
In light of the large training times of conventional deep learning models, this short time window encourages the use of efficient techniques, particularly the use of pre-trained models. The performance measurement starts with the presentation of the training data (and the inputs of the test split), and the AutoDL system can train in increments and interleave test predictions; however, the time for making predictions also counts as part of the run-time budget. Consequently, AutoDL systems need to decide when to make predictions to improve performance. 

For a formal description of the metric, as well as an example of a learning curve plot under the competition metric, please see Appendix \ref{appendix:alc_metric}.

\subsection{Benchmarks and Training Protocol}
Overall, we evaluate our ZAP methods under two benchmarks: one based on the ZAP meta-dataset which we refer to as \emph{ZAP benchmark} and the original AutoDL benchmark. For the AutoDL benchmark, we submit our ZAP methods trained on the ZAP meta-dataset. The following describes how we train and evaluate our methods on the ZAP benchmark. We evaluate in a ``leave-one-core-dataset-out protocol'' that avoids any possibility of an information leak across augmented datasets. Specifically, we meta-train our methods on 34 out of the original 35 datasets, plus their augmented versions, and test on the held-out original dataset. We average the resulting performance over 35 outer loop iterations holding out a different core dataset each time. We further apply 5-fold inner cross-validation to optimally identify the best stopping epoch while monitoring validation performance.
We evaluate each method (including the baselines) 10 times with different seeds and report averages, standard deviations, boxplots, and statistical tests over these 10 results.

For evaluation under the AutoDL benchmark, we upload our ZAP methods trained on the ZAP meta-dataset as well as the baselines to the submission board, made available to us by the challenge organizers. We report the performances on the five undisclosed final datasets of different domains (objects, aerial, people, medical, handwriting recognition) across 10 submissions. Here, we used the same hyperparameters from the ZAP benchmark.

\subsection{Baselines}
\label{exp: baselines}
To assess the performance of our proposed method, we compare it against multiple baselines, which we describe here. Aside from a random selection of one of our 525 carefully designed DL pipelines, and the single best pipeline on average across the datasets, we chose the top-3 winners of the 2019 ChaLearn AutoDL Challenge: DeepWisdom, DeepBlueAI, and PASA NJU.

Given a novel dataset and our 525 selected DL pipelines, \textbf{random selection} uniformly samples one of these pipelines and \textbf{single-best} picks the one which performs best on average over $\mathcal{D}$. We average the random selection baseline over three random selections.

\begin{table}[ht]
    \small
    \caption[Table caption]{\textbf{Summary of winner techniques} All contenders use ImageNet pre-trained networks and FAA denotes Fast AutoAugment.}
    \begin{center}
    \begin{tabular}{l l p{28mm}}
        \toprule
        \textbf{Solution} & \textbf{Augmentation} &\textbf{ML technique} \\
        \midrule
        DeepWisdom  & FAA & \scriptsize ResNet18/9 \newline Meta-trained solution agents \\
        DeepBlueAI & FAA & \scriptsize ResNet18 \newline Adaptive ensemble learning \\
        PASA NJU & Simple & \scriptsize ResNet18/SeResnext50 \newline Data adaptive preprocessing \\
        \bottomrule
    \end{tabular}
    \end{center}
    \label{tab:winner_baselines}
\end{table}

The challenge winner baselines build their methods on top of AutoCLINT~\cite{baek-arxiv20a}, with the following modifications (summarized in Table \ref{tab:winner_baselines}):
\begin{itemize}
\vspace*{-0.2cm}
    \item \textbf{DeepWisdom} initially caches mini-batches with a pre-trained ResNet-18 model and quickly switches to a pre-trained ResNet-9 by inputting cached batches first. They initialize the networks with ImageNet pre-trained parameters except for the batch normalization and bias layers. After an initial optimization phase, they turn on Fast AutoAugment~\cite{lim-neurips19a}.
\vspace*{-0.2cm}
    \item \textbf{DeepBlueAI} initializes a pre-trained ResNet-18 network and adapts some of the model hyperparameters (image size, steps per epoch, epoch after which starting validating and fusing results, etc.) to the target dataset. They also ensemble the latest $n$ predictions to stabilize them. Later in the procedure, they augment the dataset by Fast AutoAugment for a limited budget.
\vspace*{-0.2cm}
    \item \textbf{PASA NJU} pre-processes the data with a data-adaptive strategy by first sampling images to analyze. Then they crop images to a standard shape and apply image flip augmentations. They start the training with a pre-trained ResNet-18 and switch to SeResNext-50~\cite{hu-arxiv17a} when no further improvement is expected on the validation score.
\end{itemize}

\subsection{Results on the ZAP Benchmark} 
\begin{figure}[ht]
\begin{centering}

    {\includegraphics[scale=0.3]{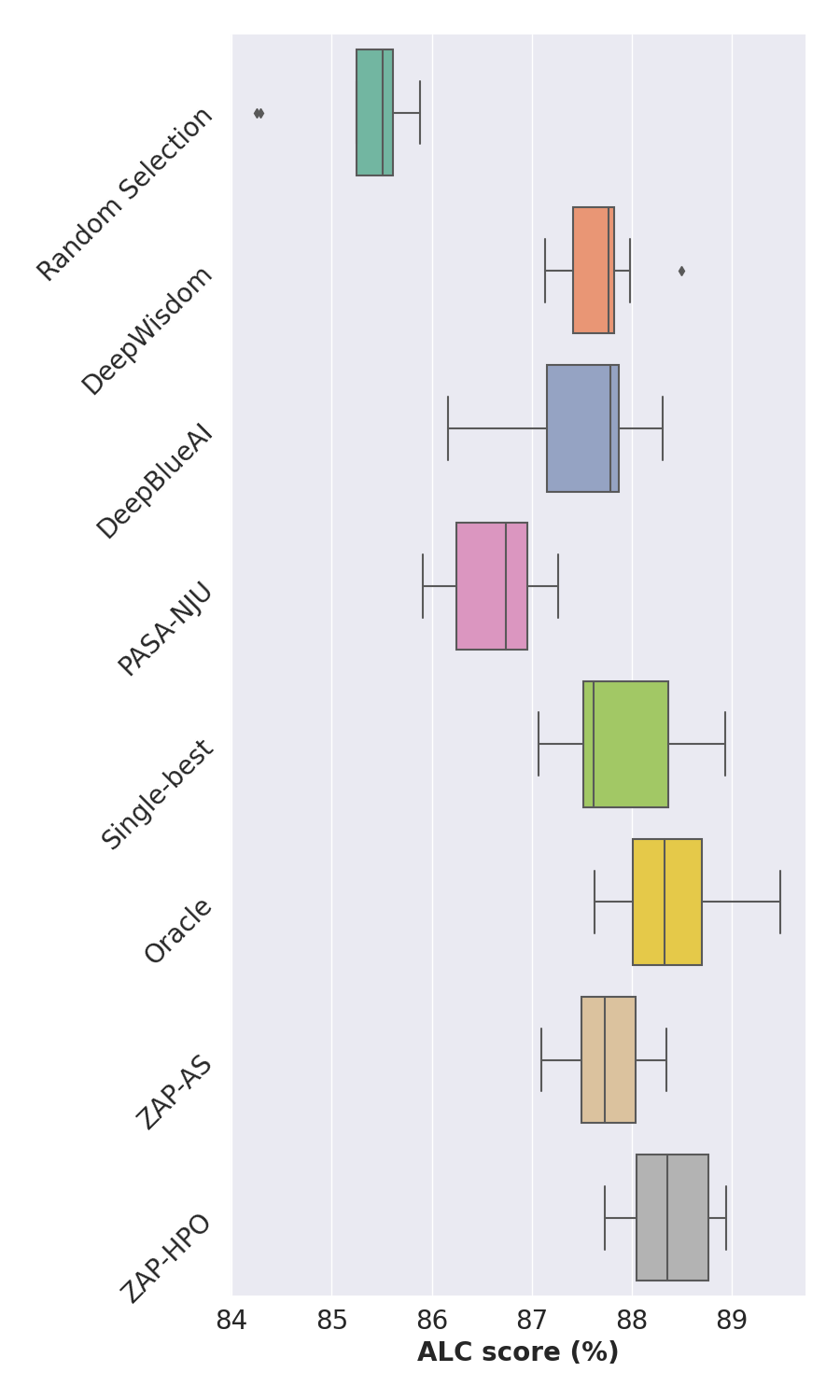}}

    \caption[]{\textbf{ALC scores of our approach vs. winner baselines} over 525 datasets and 10 repetitions. Our ZAP methods clearly improve over the challenge winners (higher is better), by almost 1 point. Our geometry-aware zero-shot HPO version of ZAP with its binary pairwise ranking objective works best.}
    \label{fig:benchmark_boxplot_best}
\end{centering}
\vspace*{-0.3cm}
\end{figure}

In Figure~\ref{fig:benchmark_boxplot_best}, we depict the performance of our ZAP methods ZAP-AS and ZAP-HPO  compared to the winner baselines of the AutoDL challenge. Our algorithm-selection-based ZAP-AS method already outperforms the competition winners, and our geometry-aware ranking-based model, ZAP-HPO, performs even significantly better.

In Table \ref{tab:ranking_simple}, we report the rank of the scores identifying the winners across the two main metrics. Our proposed method wins in both ALC score, i.e., the metric for which it was optimized, but also in terms of the normalized area under the curve (Normalized AUC). 

\begin{table}[ht]
    \small
    \caption[Table caption]{\textbf{Ranking our approach vs. winner baselines on the ZAP benchmark.} We rank the solutions per test dataset and report average ranks over the 525 datasets (averages over 10 repetitions). ZAP clearly performs best (lower is better), both in terms of ALC (which it was optimized for) and also in terms of Normalized AUC.}
    \begin{center}
    \begin{tabular}{l c c}
        \toprule
        \textbf{Solution} & \textbf{Rank (ALC)} & \textbf{Rank(Normalized AUC)}\\
        \midrule
        DeepWisdom & $2.53\pm0.06$ & $2.63\pm0.03$ \\
        DeepBlueAI & $2.64\pm0.04$ & $2.73\pm0.03$ \\
        PASA NJU & $2.76\pm0.03$ & $2.56\pm0.03$\\
        ZAP-HPO & $\mathbf{2.07\pm0.05}$ & $\mathbf{2.08\pm0.03}$\\
        \bottomrule
    \end{tabular}
    \end{center}
    \label{tab:ranking_simple}
\end{table}

\subsection{Results for a Sparsely-filled Cost Matrix}
\begin{figure}[ht]
\begin{centering}

    {\includegraphics[scale=0.3]{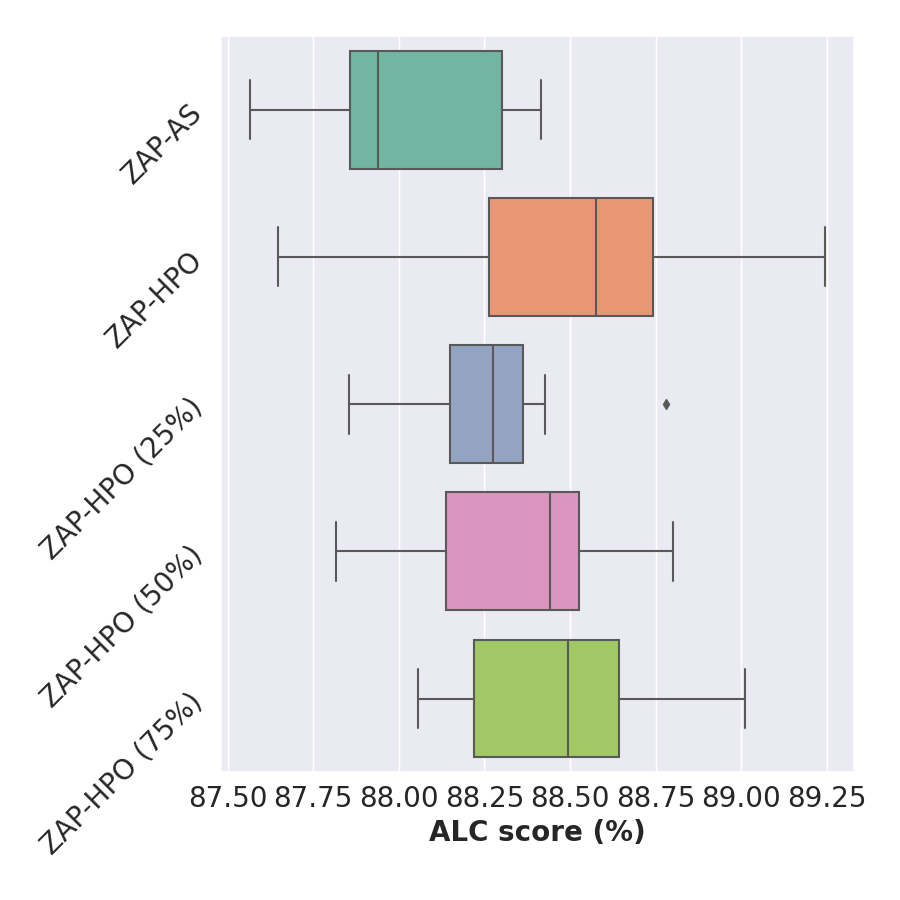}}
\vspace*{-0.6cm}
    \caption[]{\textbf{ALC scores of ZAP-HPO when meta-trained on a full (dense) vs. sparse cost matrix} over 525 datasets and 10 repetitions. The density of the matrices are denoted as 75\%, 50\%, and 25\%. }
    \label{fig:benchmark_boxplot_sparse}
\end{centering}
\end{figure}

To further investigate the source of the gains in our model, we propose a more realistic setting, where the cost matrix includes missing values. While algorithm selection methods, such as ZAP-AS, require the dense cost matrix, our geometry-aware rank-based ZAP-HPO method handled missing values gracefully.
To evaluate this quantitatively, next to ZAP-AS and ZAP-HPO with the full cost matrix, we also evaluate ZAP-HPO based on cost matrices that only have 75\%, 50\%, and 25\% of the entries (dropped at random) remaining.
As Figure \ref{fig:benchmark_boxplot_sparse} shows, ZAP-HPO's performance loss due to missing entries is quite small, and even with only 25\% remaining entries, it still performs similarly to ZAP-AS. We believe that ZAP-HPO's low sensitivity to missing values stems from the capacity of the model to capture the correlation across the pipelines in the geometric space. It can hereby generalize well and "impute" missing values of the cost matrix. Furthermore, as shown in Table~\ref{tab:ranking_sparse}, even with missing values in the cost matrix, ZAP-HPO clearly outperforms the winners of the AutoDL competition. 

\begin{table}[ht]
\small 
\caption[]{\textbf{The ranking of ZAP-HPO when optimized with a sparse cost matrix} shows we still outperform the AutoDL challenge winner solutions.}
    \begin{center}
    \begin{tabular}{l c c c}
        \toprule
        \textbf{Solution} & \textbf{75\%} filled & \textbf{50\%} filled & \textbf{25\%}  filled\\
        \midrule
        DeepWisdom & \small$2.52\pm0.06$ & \small$2.52\pm0.05$ & \small$2.52\pm0.05$\\
        DeepBlueAI & \small$2.64\pm0.05$ & \small$2.64\pm0.04$ & \small$2.63\pm0.04$\\
        PASA NJU & \small$2.75\pm0.03$ & \small$2.75\pm0.04$ & \small$2.75\pm0.04$\\
        ZAP-HPO (sparse) & \small$\mathbf{2.09\pm0.06}$ & \small$\mathbf{2.09\pm0.05}$ & \small$\mathbf{2.10\pm0.04}$\\
        \bottomrule
    \end{tabular}
    \end{center}
    \label{tab:ranking_sparse}
    \vspace*{-0.3cm}
\end{table}

\subsection{Results on the AutoDL Benchmark}
In Table \ref{tab:ranking_dashboard} we report the average rank performances for the AutoDL benchmark on the five undisclosed datasets across 10 submissions.
We highlight that, unlike the winner baselines, we did not use the challenge feedback datasets to optimize the base model and zero-shot model hyperparameters for the final submission but reused (only) the ones from the ZAP benchmark. Due to a difference in distributions between these benchmarks, it cannot be taken for granted that our method generalizes to the AutoDL competition datasets. However, as Table \ref{tab:ranking_dashboard} shows, ZAP-HPO clearly outperforms the winner baselines on this AutoDL setting. It also clearly outperforms the single-best and random baselines (which have ranks $ 2.7 (\pm 0.1)$ and $3.42 (\pm 0.68)$, respectively). In this setting of generalizing out of distribution, the more conservative ZAP-AS method performs even slightly better than ZAP-HPO, with average ranks of $1.81 (\pm 0.3)$ vs. $2.16 (\pm 0.15$).
\begin{table}[h!]
    \caption[Table caption]{\textbf{Ranking our approach vs. winner baselines on the AutoDL benchmark.} We rank the solutions per test dataset and report average ranks over the five AutoDL benchmark final datasets (averages over 10 submissions).}
    \begin{center}
    \begin{tabular}{l c}
        \toprule
        \textbf{Solution} & \textbf{Rank (ALC)} \\
        \midrule
        DeepWisdom & $2.46\pm0.13$ \\
        DeepBlueAI & $2.76\pm0.08$ \\
        PASA NJU & $2.62\pm0.11$\\
        ZAP-HPO & $\mathbf{2.16\pm0.15}$\\
        \bottomrule
    \end{tabular}
    \end{center}
    
    \label{tab:ranking_dashboard}
\end{table}

\section{Conclusion}
\label{sec:conclusion}

In this paper we extend the realm of AutoML to address the common problem of fine-tuning pre-trained Deep Learning (DL) models on new image classification datasets. Concretely, we focus on deciding which pre-trained model to use and how to set the many hyperparameters of the fine-tuning procedure, in a regime where strong anytime performance is essential. We formalize the problem as Zero-shot AutoML with Pre-trained Models (ZAP), which transfers knowledge from a meta-dataset of a number of DL pipelines evaluated on a set of image datasets. In that context, we open-source the largest meta-dataset of evaluations for fine-tuning DL pipelines with 275K evaluated pipelines on 35 popular image datasets (2871 GPU days of compute). Furthermore, we propose two approaches for tackling ZAP: (i) formulating it as an instance of the algorithm selection (AS) problem and using AS methods, and (ii) a novel zero-shot hyperparameter optimization method trained with a ranking objective. Our methods clearly achieve the new state of the art in terms of anytime Automated Deep Learning (AutoDL) performance and significantly outperform all the solutions of the 2019 ChaLearn AutoDL Challenge.  

\section{Limitations}
As mentioned before, computing the cost matrix is expensive. In particular, when training deep models, early optimization phases are often noisy and the final performance of a model is difficult to predict. Consequently, the accuracy of the cost matrix is directly coupled to the duration of the deep model training. Applying early-stopping methods to reduce the expenses of determining a cost matrix is thus challenging. Another limitation we observed is the sensitivity to the zero shot model's hyperparameters. For example, we noticed that using a least-squares or a triplet margin objective performed significantly worse than our ZAP-HPO objective. Lastly, it is not clear a-priori which attributes best describe the DL pipelines in order to achieve the best performance and selecting a sub-optimal set may lead to a deterioration of performance.
\section*{Acknowledgements}
This paper builds on and extends our original submission to the AutoDL challenge, and we are indebted to everyone who helped with that submission; in particular, we would like to thank Danny Stoll for his help on the initial hyperparameter configuration space design and data augmentation as well as Marius Lindauer for his help with AutoFolio. We also thank Dipti Sengupta for her help in reviewing both code and the paper. Moreover, we acknowledge funding by Robert Bosch GmbH, by the Deutsche Forschungsgemeinschaft (DFG, German Research Foundation) under grant number 417962828, by the state of Baden-W\"{u}rttemberg through bwHPC, and the German Research Foundation (DFG) through grant no INST 39/963-1 FUGG, by TAILOR, a project funded by the EU Horizon 2020 research, and innovation programme under GA No 952215, and by European Research Council (ERC) Consolidator Grant ``Deep Learning 2.0'' (grant no.\ 101045765). Funded by the European Union. Views and opinions expressed are however those of the author(s) only and do not necessarily reflect those of the European Union or the ERC. Neither the European Union nor the ERC can be held responsible for them.

\bibliography{bib/references,bib/lib,bib/shortproc,bib/shortstrings}
\bibliographystyle{icml2022}

\newpage
\appendix
\onecolumn
\section{Appendix}
\subsection{Search Space of DL Pipelines}
\label{appendix:dl_search_space}
We list the search space of the DL pipelines clustered into two groups. The first group are general DL hyperparameters including the architecture and fine-tuning strategy. The second group defines the early stopping, validation and test strategies during the 20 minutes of training, including set sizes and evaluation timings.
\vspace{-3pt}
\begin{table}[ht!]
    \caption[Table caption]{\textbf{The search space of our DL pipelines} clustered into two groups. The first group are general DL hyperparameters including the architecture and fine-tuning strategy. The second group defines the early stopping, validation and test strategies during the 20 minutes of training, including set sizes and evaluation timings.} 
\begin{center}
    \begin{tabular}{lcc}
        \toprule
        \textbf{Name} & \textbf{Type, Scale} & \textbf{Range} \\ 
        \midrule
        Batch size & int, log & $[16, 64]$ \\ 
        Learning rate & float, log & $[10^{-5}, 10^{-1}]$ \\
        Min learn. rate & float, log & $[10^{-8}, 10^{-5}]$ \\
        Weight decay & float, log & $[10^{-5}, 10^{-2}]$ \\
        Momentum & float & $[0.01, 0.99]$ \\
        Optimizer & cat & \{SGD, Adam,  \\
         &  & \text{   }AdamW\} \\
        Nesterov & cat & \{true, false\} \\
        Amsgrad & cat & \{true, false\} \\
        Scheduler & cat & \{plateau, cosine\} \\
        Freeze portion & cat & \{$0.0,0.1,\dots,0.5$\}\\
        Warm-up mult. & cat & \{$1.0$, $1.5$, $\dots$, $3.0$\}\\
        Warm-up epoch & int & $[3, 6]$ \\
        Architecture & cat & \{ResNet18, EffNet-b0 \\
         &  & \text{   }EffNet-b1, EffNet-b2\} \\
        \hline
        Steps per epoch     & int, log & $[5, 250]$ \\
        Early epoch         & int& $[1, 3]$ \\
        CV ratio      & float & $[0.05, 0.2]$ \\
        Max valid count     & int, log & $[128, 512]$ \\
        Skip valid thresh. & float & $[0.7, 0.95]$ \\
        Test freq. & int & $[1, 3]$ \\
        Test freq. max & int & $[60, 120]$ \\
        Test freq. step & int & $[2, 10]$ \\
        Max inner loop & float & $[0.1, 0.3]$ \\
        \# init samples   & int,log & $[128, 512]$ \\
        Max input size            & int & $[5, 7]$ \\
        $1^{\text{st}}$ simple model & cat & \{true, false\} \\
        Simple model & cat & \{SVC, NuSVC, RF, LR\} \\
        \bottomrule
    \end{tabular}
    \end{center}
    \label{tab:ss_autodl1}
\end{table}

\subsection{Dataset Augmentation}
\label{appendix:dataset_augmentation}
\begin{figure}[ht]
\begin{centering}

    {\includegraphics[scale=0.5]{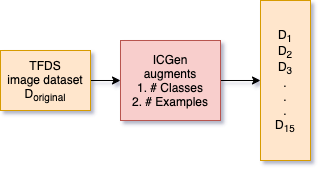}}

    \caption[]{\textbf{Dataset augmentation flow}} 
    \label{fig:chart_icgen}
\end{centering}
\end{figure}

\begin{figure}[ht]
\centering
    \includegraphics[scale=0.27]{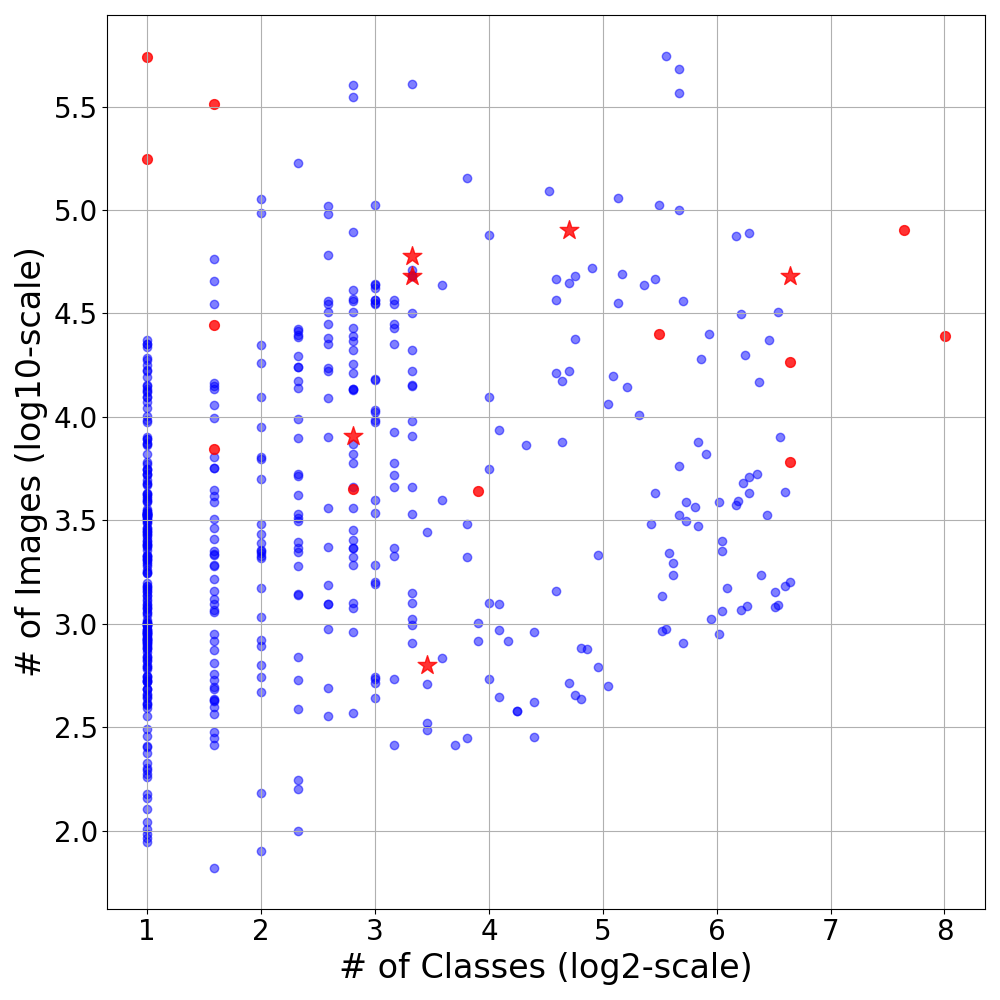}
    \caption[]{\textbf{Distribution of the meta-features} where each point corresponds to a dataset. Blue points come from our meta-dataset, whilst red ones are the datasets provided by AutoDL challenge. Star and point markers are public AutoDL datasets and private AutoDL datasets (from feedback and final phases), respectively.}
    \label{fig:scatterplot_metafeatures}
\end{figure}

In the following Table \ref{tab:dataset_domains} we list all TFDS datasets we used for creating our 525 datasets with their respective domains. For each dataset $\mathcal{D}_{original}$ we create 15 subsets by sampling number of classes from range $[2, 100]$ and min/max number of samples per class from range $[20, 10^5]$ such that $\mathcal{D}_i \subseteq \mathcal{D}_{original},\forall i \in \{1, \dots, 15\}$ (Figure \ref{fig:chart_icgen}). Remark that there is no procedure on sample level, meaning that subsets inherit image resolutions and channels from the original dataset as given in the Tables \ref{tab:mf_stats_1} and \ref{tab:mf_stats_2}. Figure \ref{fig:scatterplot_metafeatures} contains the number of sample and class distributions of these subsets along with the AutoDL Challenge benchmark datasets. 

\begin{table}[ht]
    \caption[Table caption]{\textbf{Domains of the original datasets} }
    \begin{center}
    \begin{tabular}{l p{120mm}}
        \hline
        \textbf{Domain} & \textbf{Datasets}\\
        \hline
        Objects & Cifar100~\cite{cifar}, Cifar10, Horses or Humans~\cite{Moroney2019_HorsesHumans}, CycleGAN Horse2zebra~\cite{Zhu2017_CycleGAN}, CycleGAN Facades, CycleGAN Apple2orange, Imagenette~\cite{Howard_Imaginette}, Coil100~\cite{Nene1996_Coil100}, Stanford Dogs~\cite{Khosla2011_StanfordDogs}, Rock, Paper and Scissors~\cite{Moroney2019_RPS}, TF Flowers~\cite{TF_2019_Flowers}, Cassava~\cite{Mwebaze2019_Cassava}, Fashion MNIST~\cite{xiao-corr17a}, Cars196~\cite{Krause2013_Cars196}, Cats vs Dogs~\cite{Elson2007_CatsDogs}, ImageNet Resized 32x32~\cite{Chrabaszcz2017_ImageNetDownsampled} \\
        Characters & Cmaterdb Devanagari~\cite{Das2012_Cmaterdb2, Das2012_Cmaterdb}, Cmaterdb Bangla, MNIST\cite{mnist}, KMNIST~\cite{Clanuwat2018_Kmnist}, EMNIST Byclass~\cite{Cohen2017_Emnist}, EMNIST MNIST, Cmaterdb Telugu, EMNIST Balanced, Omniglot~\cite{Lake2015_Omniglot}, SVHN Cropped~\cite{Netzer2011_SVHN} \\
        Medical & Colorectal Histology~\cite{colorectalhistology}, Malaria~\cite{Rajaraman2018_Malaria} \\
        Aerial & Uc Merced~\cite{Nilsback08_UC}, CycleGAN Maps, Eurosat RGB~\cite{eurosat} \\
        Drawings/Pictures & CycleGAN Vangogh2photo, CycleGAN Ukiyoe2photo \\
        \hline
    \end{tabular}
    
    \end{center}
    \label{tab:dataset_domains}
\end{table}
\begin{table}[b]
    \caption[Table caption]{\textbf{Resolution Distribution} of datasets}
    \begin{center}
    \begin{tabular}{l c}
        \hline
        \textbf{Resolution} & \textbf{\# of Datasets}\\
        \hline
        $28\times28$ & 90 \\
        $32\times32$ & 105 \\
        $64\times64$ & 15 \\
        $105\times105$ & 15 \\
        $128\times128$ & 15 \\
        $150\times150$ & 15 \\
        $256\times256$ & 90 \\
        $300\times300$ & 30 \\
        $600\times600$ & 15\\
        Varying & 135 \\
        \hline
    \end{tabular}
    
    \end{center}
    \label{tab:mf_stats_1}
\end{table}

\begin{table}
    \caption[Table caption]{\textbf{Image channels} distribution of datasets}
    \begin{center}
    \begin{tabular}{l c}
        \hline
        \textbf{Channel} & \textbf{\# of Datasets}\\
        \hline
        B\&W & 90 \\
        RGB & 435 \\
        \hline
    \end{tabular}
        \end{center}
    \label{tab:mf_stats_2}
\end{table}



\subsection{AutoDL: Area Under the Learning Curve (ALC) Metric}
\label{appendix:alc_metric}


In the 2019 ChaLearn AutoDL challenge and also in all our experiments, the main performance metric is the Area under the Learning Curve (ALC). Formally, we test the currently trained model on a test set $p_i$ at time $t_i$ by calculating a scalar score, the Normalized Area Under ROC Curve (AUC):
\begin{equation}
    s_i = 2*AUC(\Vec{p_i})-1
\end{equation}

We then convert this score to a time-sensitive step function
\begin{equation}
    s(t) = step\_fn(\Vec{s})
\end{equation}

and we also transform the time non-linearly between $[0, 1]$ such that the performance on the first minute is weighted roughly the same as the last 10 minutes of the budget:

\begin{equation}
    \tilde{t}(t) = \frac{\log{(1+t/t_0)}}{\log{(1+T/t_0)}}
\end{equation}

where $T = 1200$ is the total default training budget and $t_0 = 60$ is the default reference time. 

Finally, we measure the ALC by:

\begin{equation}
ALC =  \int_{0}^{1} s(t) d\tilde{t}(t)
\end{equation}

\begin{figure}
\centering
    \includegraphics[scale=0.5]{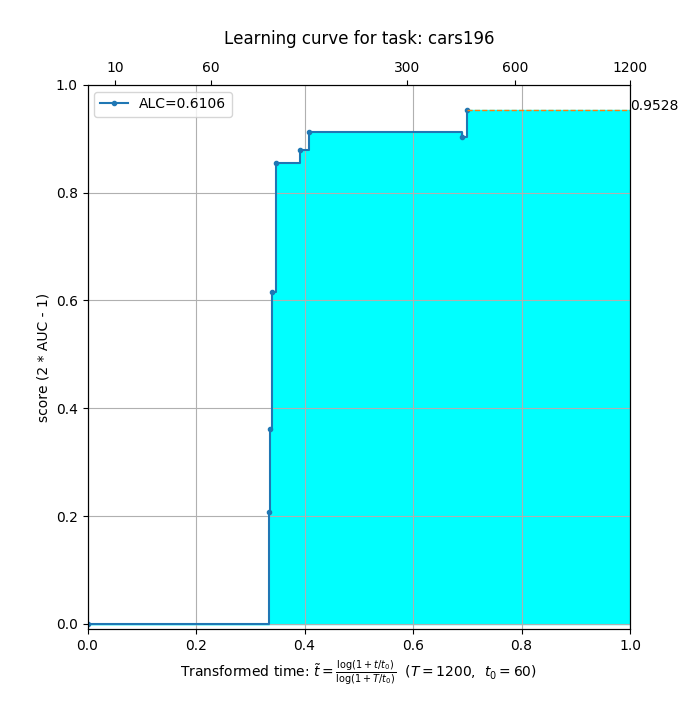}
    \caption{\textbf{Learning curve of a task} is the step function of normalized AUC (NAUC) scores received during the 20 minutes and the light blue area underneath is the area under the learning curve(ALC). Every dark blue point (steps) corresponds to a set of predictions made and y-axis to its NAUC score. When the procedure stops early, the final NAUC is interpolated to the end with a horizontal line. Top x-axis is the time limited to $1200$ seconds and the bottom axis is the transformed version between $[0,1]$. Visualization clearly shows that the first $60$ seconds contributes to more than $20\%$ of the total score and has the same weight as the last $600$ seconds. Remark that total budget $T=1200$ and reference time constant $t_0=60$ here, the same values as in our experiments.}
    \label{fig:alc_curves}
\end{figure}

For more details on the metric, we refer the reader to \cite{liu-tpami21a}. We depict an example of an ALC plot in Figure \ref{fig:alc_curves}.

\subsection{Hardware Setup}
\label{appendix:hw_setup}
Due to the anytime performance measurements of the AutoDL challenge's training and evaluation protocol and the resulting importance on wall clock time, we ensured that all experiments in this work were run on machines with the same hardware setup. The specification of our machines is the following: AMD EPYC 7502 32-Core Processor, NVIDIA GeForce RTX 2080 Ti, 500GB RAM, CUDA version 11.5, Ubuntu 20.04.3 LTS.


\end{document}